\documentclass[runningheads]{llncs}

\usepackage{eccv}

\usepackage{eccvabbrv}

\usepackage{graphicx}
\usepackage{booktabs}

\usepackage[accsupp]{axessibility}  

\usepackage{kotex}
\usepackage{moresize}
\usepackage{array}

\usepackage[pagebackref,breaklinks,colorlinks,citecolor=eccvblue]{hyperref}

\usepackage{orcidlink}

\usepackage{amsmath}
\usepackage{amssymb}
\usepackage{array}

\begin{document}

\title{Ground-A-Score: Scaling Up the Score Distillation for Multi-Attribute Editing}

\titlerunning{Ground-A-Score}

\author{Hangeol Chang\inst{*} \and
Jinho Chang\inst{*} \and
Jong Chul Ye\inst{}}

\authorrunning{Hangeol Chang, Jinho Chang, and Jong Chul Ye.}

\institute{Kim Jaechul Graduate School of AI, KAIST  \\
\email{\{hangeol, jinhojsk515, jong.ye\}@kaist.ac.kr}\\
* Equal contribution}
\maketitle
\begin{center}
	\centering
 \captionsetup{type=figure}
 \includegraphics[width=\textwidth]{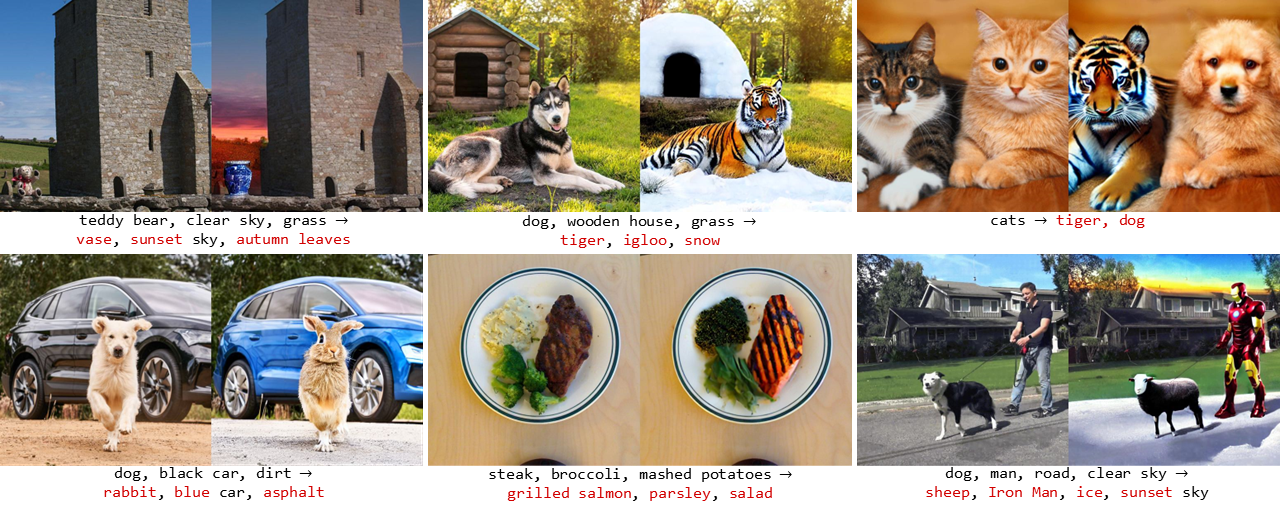}
	\caption{
Multi-attribute image editing results by Ground-A-Score.
}
	\label{fig_first}
\end{center}

\begin{abstract}
Despite recent advancements in text-to-image diffusion models facilitating various image editing techniques, complex text prompts often lead to an oversight of some requests due to a bottleneck in processing text information. 
To tackle this challenge, we present {\em Ground-A-Score}, a simple yet powerful model-agnostic  image editing method 
by incorporating grounding during score distillation. This approach ensures a precise reflection of intricate prompt requirements in the editing outcomes,
taking into account the prior knowledge of the object locations within the image.
Moreover, the selective application with a new penalty coefficient and contrastive loss helps to precisely target editing areas while preserving the integrity of the objects in the source image. 
 Both qualitative assessments and quantitative analyses confirm that Ground-A-Score successfully adheres to the intricate details of extended and multifaceted prompts, ensuring high-quality outcomes that respect the original image attributes.\
\end{abstract}

\section{Introduction}
\label{sec:intro}
With the recent advances in generative models for diverse data domains~\cite{gpt3, audio_diffusion, video_diffusion} including images and texts, various Text-to-Image (T2I) models~\cite{unidiffuser, d3pm_i2t} have been proposed to synthesize and edit images according to text prompts. Particularly, diffusion models~\cite{ddpm, score_sde, ddim, d3pm} have demonstrated remarkable achievement, 
 generating high-quality images with high fidelity against the input text condition.

Leveraging the informative text-conditioned image distribution learned from these recent T2I diffusion models, various studies have been ongoing to utilize large T2I models for image modification~\cite{prompt_to_prompt, sdedit, plug_and_play, textualinversion, masactrl}. Among them, distillation-based methods~\cite{sds, dds} are  simple yet powerful model-agnostic  approaches that
optimize the original image directly using the diffusion model's prior text-to-image generation knowledge. These approaches effectively achieve gradual, desired changes while preserving the original image's structural appearance or unrelated background.

Unfortunately, the score distillation approaches often fail when the user demands to edit multiple features simultaneously.
This phenomenon mainly comes from the inherited limitation from the base T2I models, which often omit specific objects or incorrectly compose the components when generating multiple instances at once~\cite{compositional_score, rpg}. 
This phenomenon happens more often when the editing target is small or located on the edges as will be more detailed in Section~\ref{method_reg}. 

\begin{figure}[tb!]
	\centering
 \includegraphics[width=\textwidth]{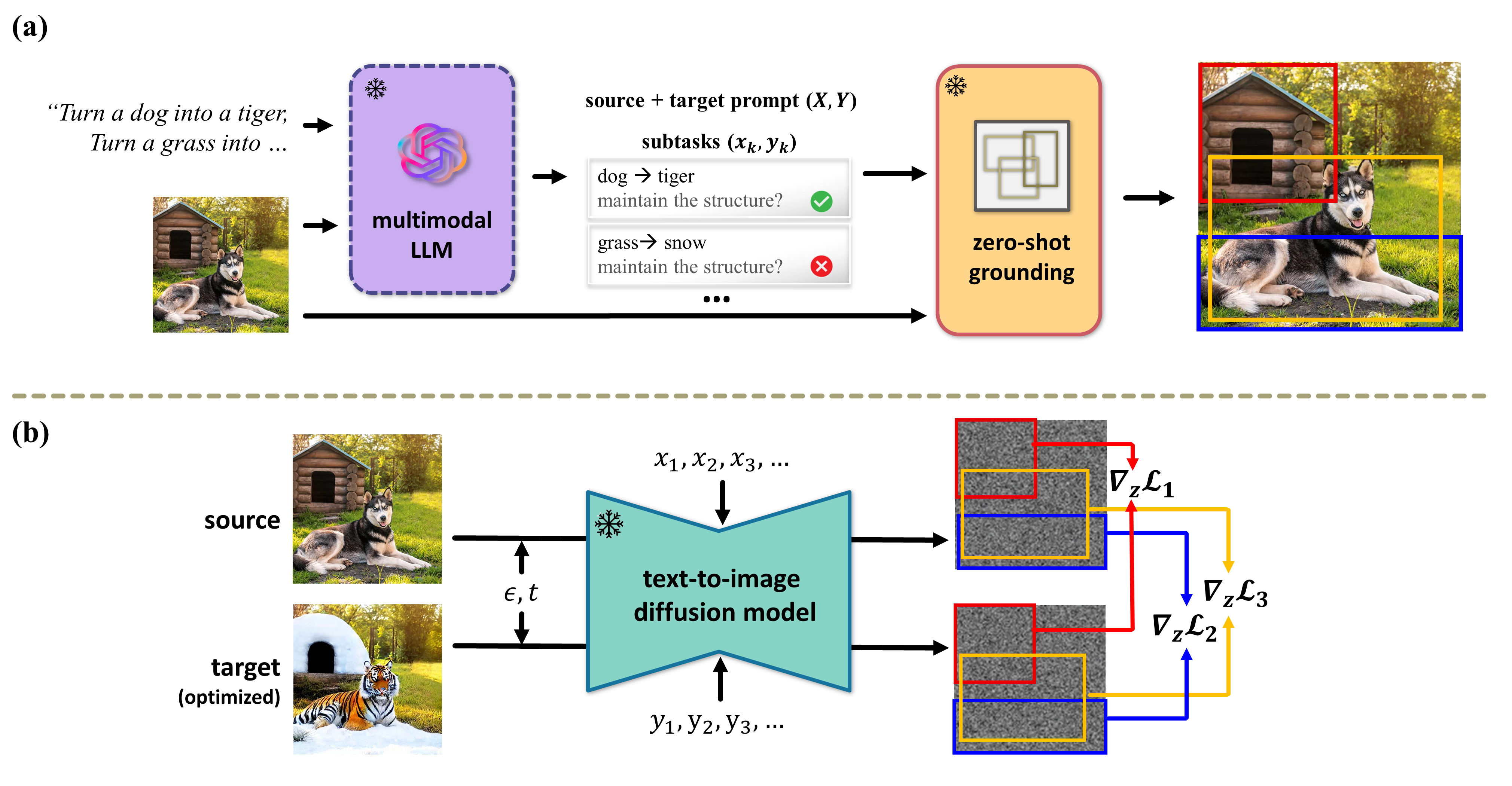}
	\caption{
The overview of the proposed pipeline for image editing with complex user requests. (a) We leverage the prior knowledge from the multimodal LLM and the zero-shot grounding model to break down the user request into multiple image editing subtasks for a single entity. (b) A pre-trained text-to-image diffusion model is used for each subtask to obtain a corresponding gradient for the source image. These gradients are masked and aggregated to get a total gradient that is efficient and stable.
}
	\label{fig_overview}
\end{figure}

To address this, here we propose an image editing pipeline called Ground-A-Score, that can overcome the aforementioned issues in the conventional score-distillation methods.
Similar to the works that attempt to uphold the T2I models' image generation performance into more complex prompts~\cite{compositional_attention1,boxdiff,rpg}, Ground-A-Score breaks down a complex editing prompt into multiple individual modification subtasks. However, in contrast to these approaches that require extensive feature engineering or modification of the internal structure of diffusion models,
 Ground-A-Score performs the subtasks in a model-agnostic manner,
 by simply splitting the distillation loss according to the grounded subtask.
Then, the calculated image editing gradient for each subtask is selectively aggregated with grounding information to obtain a total optimization direction.
Additionally, the occasion of object distortion was minimized by 
introducing a regularization coefficient that penalizes the undesired changes.
The resulting  Ground-A-Score pipeline is illustrated in Figure~\ref{fig_overview}.
Our contribution can be summarized as follows:
\begin{itemize}
    \item Ground-A-Score tackled long and complex editing prompts with the principle of divide-and-conquer during  the score distillation.
    \item Ground-A-Score automates the preparation of editing requirements, such as carefully designed prompts and information about object locations, via zero-shot grounding models and multi-modal LLMs.
    \item We show that Ground-A-Score outperforms the existing image editing models, and faithfully reflects the multiple input modification queries with minimal image corruption.
\end{itemize}

\section{Related Works}
\subsubsection{Image Manipulation with Text-to-Image Diffusion Models}

With the remarkable success of diffusion models in the image generation task~\cite{ddpm, ddim, edm}, various T2I-diffusion models~\cite{imagen, stablediffusion} are suggested that can generate images that match the desired text description.
These off-the-shelf T2I diffusion models contain valuable information about text-conditioned image distribution, and often exploit interpretable spatial correlation between the sampled image and the input text via cross-attention in the score network~\cite{unet}.

The exploitation of the captured conditional distribution from these pre-trained T2I diffusion models enabled techniques for diverse tasks such as image translation~\cite{sdedit} and inverse problems~\cite{treg}. 
Especially for the image editing task where the preservation of the source image is essential, various methods are applied in the diffusion models' sampling process.
The Prompt-to-Prompt~\cite{prompt_to_prompt} and inversion methods~\cite{nulltext_inversion, fixedpointinversion,negativepromptinversion} involve identifying the source image's sampling trajectory through the model. This process replicates the model's cross-attention map to maintain the spatial structure of the original image while transforming it.  Similarly,  MasaCtrl \cite{masactrl} replicates the self-attention map to preserve the source image characteristics.

Meanwhile, DreamFusion~\cite{dds} suggests a  Score Distillation Sampling (SDS), which utilizes a pre-trained T2I diffusion model but is independent of its sampling process. 
By optimizing a parameterized image generator to minimize the error of the pre-trained T2I diffusion model, SDS was able to generate images and neural radiance fields that align with the given text descriptions. 
However, the image gradient from SDS has an inconsistency from the assumption that the diffusion model loss is zero when the image and text are matched, which leads to the blurring of the background when it's applied to image editing. Delta Denoising Score (DDS)~\cite{dds} addressed this issue by subtracting the SDS loss between the input image and matching text from the baseline SDS loss. 
Nam \etal ~\cite{cds} reported that a Contrastive Unpaired Translation (CUT) loss~\cite{cut} can be attached on top of the DDS, which helps the output image preserve the structure of the source image.
Although these distillation-based approaches effectively achieve gradual, desired changes while preserving the original image's structural appearance,
they often fail to meet all given requirements when prompts are too long or require multiple objects to generate.

\subsubsection{Improving Fidelity of Text-to-Image Diffusion models}

Several works applied a divide-and-conquer principle to separate a complex text prompt into small components, to ease the load of the T2I diffusion model and improve the output image fidelity with the given prompt.
Recent work~\cite{compositional_score} proposed the conjunction of the individual score output with multiple sub-prompts in the sampling process to generate the image with the composite prompt.
Other methods~\cite{compositional_attention1, compositional_attention2, boxdiff} tried to manipulate the cross-attention map between the prompt and the intermediate sampling image to ensure certain objects appear or are located in a specific location. 
BoxDiff~\cite{boxdiff} and Ground-A-Video~\cite{groundavideo} utilized a given grounding information of the appearances to successfully perform multi-attribute image generation and video editing.
However, most of these approaches are designed for reverse diffusion sampling, which often 
requires the extensive structural modification and feature engineering of diffusion models.

\section{Method: Ground-A-Score}

\subsection{Aggregation of Multiple Editing Guidance}
For a given  pre-trained T2I diffusion model $\hat{\epsilon}_{\phi} (z_t,t,c)$ parameterized with $\phi$, 
SDS\cite{sds} optimizes the image latent $z$ from the initial image $\bar z$  to follow the text condition $c$ by minimizing the following loss:
\begin{align}
\mathcal{L}_{SDS}(z, t, c, \epsilon) &= w(t) \left\| \hat{\epsilon}_{\phi} (z_t,t,c) - \epsilon \right\|^2_2 \; \label{eq1}
\end{align}
where $\epsilon\sim\mathcal{N}(0,I)$ and $t\sim\mathcal{U}(0,1)$ denotes the added noise and the noise timestep, $\hat{\epsilon}_{\phi}(z_t,t,c)$ is the modeled score output, $z_t$ is a perturbed image with timestep $t$, and $w(t)$ is a certain weight function. 
When there's a generator model $\theta$ that samples $z$, it is known that this loss can be minimized by optimizing $z$ with the following gradient.
\begin{align}
\nabla_{\theta} \mathcal{L}_{\text{SDS}} = w'(t)\left( \hat{\epsilon}_{\phi} (z_t,t,c) - \epsilon \right)\frac{\partial z}{\partial\theta} \; \label{eq1_2}
\end{align}
DDS\cite{dds} improved this image gradient to edit given images directly, by additionally using the source image $\Bar{z}$ and its text description $\Bar{c}$, based on the assumption that a non-zero gradient from the matched source image-text pairs can be treated as a noisy optimization direction:
\begin{align}
\nabla_{z} \mathcal{L}_{\text{DDS}} = w'(t)\left( \hat{\epsilon}_{\phi} (z_t,t,c) - \hat{\epsilon}_{\phi} (\Bar{z}_t,t,\Bar{c}) \right) \; \label{eq3}
\end{align}
If the T2I diffusion model utilizes a Classifier-Free Guidance (CFG) \cite{cfg} in its sampling process, the model output $\hat{\epsilon}_{\phi}(z_t,t,c)$ and $\hat{\epsilon}_{\phi}(\Bar{z}_t,t,\Bar{c})$ from \eqref{eq3} is replaced into $\hat{\epsilon}^{\omega}_{\phi} (z_t,t,c)$ and $\hat{\epsilon}^{\omega}_{\phi} (\Bar{z}_t,t,\Bar{c})$ with a CFG weight $\omega$:
\begin{align}
\hat{\epsilon}^{\omega}_{\phi} (z_t,t,c) &= (1 - \omega)\hat{\epsilon}_{\phi} (z_t,t,c) + \omega\hat{\epsilon}_{\phi} (z_t,t, \varnothing) \; \label{eq4} \\
\hat{\epsilon}^{\omega}_{\phi} (\Bar{z}_t,t,\Bar{c}) &= (1 - \omega)\hat{\epsilon}_{\phi} (\Bar{z}_t,t,\Bar{c}) + \omega\hat{\epsilon}_{\phi} (\Bar{z}_t,t, \varnothing) \; \label{eq5}
\end{align}
It is known that CUT loss can be additionally applied when structure preservation is needed, which we utilize selectively depending on the editing target (more details in Section~\ref{method_id}).

While DDS sampling effectively transforms images, it is often problematic to thoroughly optimize the image when multiple attributes need to be modified simultaneously. 
Part of this challenge comes from the initial limitations of the employed T2I diffusion model\cite{compositional_attention2}: as the text prompt gets more complex, the text encoder in the T2I diffusion model becomes an information bottleneck and fails to fully embed the contents in the prompt~\cite{compositional_score}.

To overcome this issue, we break down a complex editing prompt into multiple individual modification subtasks to calculate the score gradient on the image latent separately. 
Specifically, let us denote the sentence that represents the original image as the source prompt by $X$ and the sentence that represents the image to be transformed as the target prompt by $Y$. If $X$ and $Y$ have $n$ number of disagreements in them, we write the series of words in $X$ and $Y$ which correspond to those differences as $\{x_1, x_2, \ldots, x_n\}$ and $\{y_1, y_2, \ldots, y_n\}$, respectively. Then, we calculate a DDS gradient for each subtask $(\hat{\epsilon}^{\omega}_{\phi} (z_t,t,y_k) - \hat{\epsilon}^{\omega}_{\phi} (\Bar{z}_t,t,x_k))$ separately.
This reduces the burden of the T2I diffusion model to provide a more accurate gradient for each subtask, and these gradients can be aggregated to build a total image gradient to modify the images with all desired changes:
\begin{align}
\nabla_z \mathcal{L}_{sum} &= \sum_{k=1}^{n} \left(\hat{\epsilon}^{\omega}_{\phi} (z_t,t,y_k) - \hat{\epsilon}^{\omega}_{\phi} (\Bar{z}_t,t,x_k)\right)   \; \label{eq6_1}
\end{align}

However, a naive summation of all gradients like \eqref{eq6_1} makes the resulting image blurred, and some objects are corrupted or erased (see Section~\ref{result_abl}). This phenomenon occurs because the computed DDS gradient is not zero even in the regions unrelated to the editing in the real-world scenario. These inconsistent and noisy gradients corrupt the background and the other objects, and this effect is amplified when we sum them up with the number of the total subtasks. 

One simple regularization would be to utilize a binary mask $m_k$ which represents the region of $\Bar{z}$ corresponding to $x_k$, and prune the gradient corresponds to the outside of $m_k$. Nonetheless, as we modify multiple instances whose corresponding masks hugely overlap, this masking strategy still cannot manage the disagreement between gradients from different subtasks. 
To resolve this, we introduced an additional ``full-prompt guidance'' term using the original full-prompts $(X, Y)$ with a hyperparameter $\alpha$. With the information of the full prompt that contains the relations between each subtask's target, the additional gradient term helps to organize the resulting image and improve coherence~\cite{rpg}.
In total, we define the Ground-A-Score loss as follows:
\begin{align}
\begin{aligned}
\nabla_z \mathcal{L}_{GAS} &= \sum_{k=1}^{n} \left((\hat{\epsilon}^{\omega}_{\phi} (z_t,t,y_k) - \hat{\epsilon}^{\omega}_{\phi} (\Bar{z}_t,t,x_k))\odot m_k\right)  \\
&+ \alpha(\hat{\epsilon}^{\omega}_{\phi} (z_t,t,Y) - \hat{\epsilon}^{\omega}_{\phi} (\Bar{z}_t,t,X))\odot M   \; \label{eq6}
\end{aligned}
\end{align}
where $M$ refers to a mask representing any region where at least one of ${x_1, x_2, \ldots, x_n}$ exists (\ie $M:=m_1 \cup m_2 \cup \ldots \cup m_n$), and $\odot$ is the element-wise multiplication.
The resulting Ground-A-Score loss has two major benefits: first, it guides the distillation gradient to where individual modification should be done.
Secondly, it regularizes the modified region to minimize the variance from the inevitable noisy DDS gradient on unrelated regions.

\subsection{Null-Text Penalty for Non-Trusted Gradients} \label{method_reg}

It is often discovered that image editing with DDS distorts the target object, or even makes it disappear. We observed that this can be explained with its loss function when we rewrite the~\eqref{eq3} as a sum of two SDS loss terms:
\begin{align}
\nabla_{z} \mathcal{L}_{\text{DDS}} &= w'(t)\{(\hat{\epsilon}^{\omega}_{\phi} (z_t,t,c)-\epsilon) - (\hat{\epsilon}^{\omega}_{\phi} (\Bar{z}_t,t,\Bar{c})-\epsilon) \} \; \label{eq7} \\
&= w'(t)(\mathcal{L}_{\text{SDS}} (z, t, c, \epsilon) - \mathcal{L}_{\text{SDS}} (\Bar{z}, t, \Bar{c}, \epsilon)) \label{eq8}
\end{align}
During the image latent $z$ being optimized, the base T2I diffusion model has a chance of failing to capture the correct guidance of $\mathcal{L}_{\text{SDS}} (z, t, c, \epsilon)$. When this happens, the total DDS loss is dominated by the second term, the negative guidance of generating $\Bar{c}$ from $\Bar{z}$, which leads to erasing the object. 

\begin{figure}[t]
	\centering
 \includegraphics[width=\textwidth]{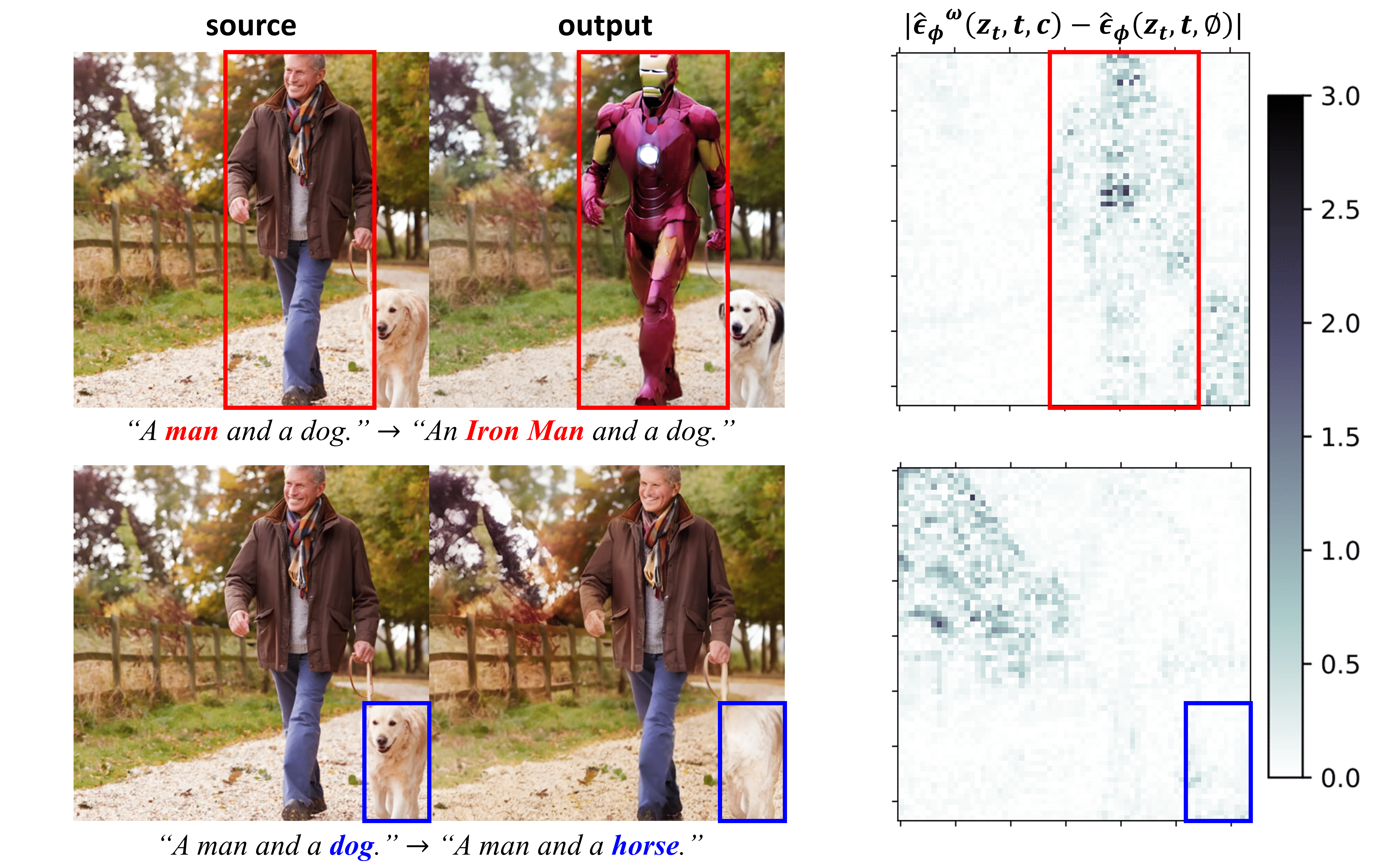}
	\caption{
The difference between the predicted noise on the target image, with the given condition and null text, in two image editing scenarios. The source image and the output from DDS are also provided. The red boxes indicate the region corresponding to the object meant to be edited.
}
	\label{fig_ntp}
\end{figure}

More specifically, we found that the predicted noise with $z$ and the target text condition $c$ is similar to the predicted noise with $z$ and the null text $\varnothing$ when this phenomenon occurs. Figure~\ref{fig_ntp} shows two scenarios of image editing with DDS, with the plot of the difference between $\hat{\epsilon}^{\omega}_{\phi} (z_t,t,c)$ and $\hat{\epsilon}_{\phi} (z_t,t,\varnothing)$ in the early stage of the editing process. When $\hat{\epsilon}^{\omega}_{\phi} (z_t,t,c)$ is distinctly different from $\hat{\epsilon}_{\phi} (z_t,t,\varnothing)$ on the object corresponds to $c$, DDS successfully converts that object to follow the input condition. On the other hand, when $|\hat{\epsilon}^{\omega}_{\phi} (z_t,t,c) - \hat{\epsilon}_{\phi} (z_t,t,\varnothing)|$ is close to 0 on the target object, the object was removed from the output image. Considering that $(\hat{\epsilon}_{\phi} (z_t,t,c) - \hat{\epsilon}_{\phi} (z_t,t,\varnothing))$ is a guidance term used in CFG, small $|\hat{\epsilon}^{\omega}_{\phi} (z_t,t,c) - \hat{\epsilon}_{\phi} (z_t,t,\varnothing)|$ in certain region implies the base T2I diffusion model treated that region is not playing a significant role to make the image agrees with the given text condition.
Although the occasion of this effect has a certain randomness throughout the iterations due to its dependence on $\epsilon$, $t$, and a moving image $z$, it happens more often when the object is small or off-centered. And as the number of editing attributes increases, there's more risk for the objects of being exposed to this problem.

Based on this analysis, we introduced a new regularizer of ``null-text penalty'' for subtasks whose $m_k$ is smaller than a certain threshold, to minimize the effect of the undesired image gradient that corrupts the object. Specifically, we modified the gradient in~\eqref{eq6} with a null-text penalty coefficient $\gamma_k\in[0, 1]$ for each subtask and multiply them for the corresponding image gradient. 
\begin{align}
\begin{aligned}
\nabla_z \mathcal{L}_{GAS} &= \sum_{k=1}^{n} \left( \gamma_k(\hat{\epsilon}^{\omega}_{\phi} (z_t,t,y_k) - \hat{\epsilon}^{\omega}_{\phi} (\Bar{z}_t,t,x_k))\odot m_k \right)  \\
&+ \alpha(\hat{\epsilon}^{\omega}_{\phi} (z_t,t,Y) - \hat{\epsilon}^{\omega}_{\phi} (\Bar{z}_t,t,X))\odot M   \; \label{eq999}
\end{aligned}
\end{align}
where 
$\gamma_k$ penalizes the non-trusted gradient from each subtask according to how $\hat{\epsilon}^{\omega}_{\phi} (z_t,t,y_k)$ is close to $\hat{\epsilon}_{\phi} (z_t,t,\varnothing)$, and is calculated as follows:
\begin{align}
\gamma_k &= \min\left(\frac{\eta}{N_{m_k}}\sum(| \hat{\epsilon}^{\omega}_{\phi} (z_t,t,y_k) - \hat{\epsilon}_{\phi} (z_t,t,\varnothing) |\odot m_k), 1\right) \; \label{eq99}
\end{align}
Here, $N_{m_k}$ is a number of non-zero elements in $m_k$, and $\eta$ is a hyperparameter that controls the range of the penalty.
When the obtained gradient occasionally guides the image to erase the object, the null-text penalty prevents the object from being deleted with small $\gamma_k$. This eventually makes the object correctly modified, which we demonstrate in more detail in Section~\ref{result_abl}.

\subsection{Automated Pipeline for Input Preparation} \label{method_id}

The proposed Ground-A-Score editing loss requires a prompt sentence $(X, Y)$ that describes the target and source image, the separated subtask pairs $(x_k, y_k)$, and their corresponding binary mask $m_k$. As the desired editing query gets more complicated, preparing these inputs manually becomes inefficient and prone to getting inaccurate prompts, especially for writing a source prompt sentence.

We have developed a user-friendly and straightforward framework for image editing, as shown in Figure~\ref{fig_overview}, integrating the aforementioned methods with a zero-shot object detection model and a multimodal Large Language Model (mLLM). When a user inputs a desired change within an image (\eg ``transform a dog into a cat''), the multimodal LLM generates the source and target image prompts $(X, Y)$ along with their respective differences $\{x_1, x_2, \ldots, x_n\}$ and $\{y_1, y_2, \ldots, y_n\}$. 
The mLLM could also determine whether an additional CUT loss~\cite{cut, cds} for structure preservation would be applied or not, using its prior knowledge about shapes and the user-given examples.
This can be done by leveraging the in-context learning capabilities of the LLM, and we feed our source image and user request queries with specific Chain-of-Thought\cite{cot} prompts. Subsequently, a zero-shot object detection model~\cite{grounding_dino} is employed to identify the mask $m_k$ corresponding to the source image component $x_k$. This information is then used to update the image with Ground-A-Score loss.

\begin{figure}[p!]
	\centering
 \includegraphics[width=\textwidth]{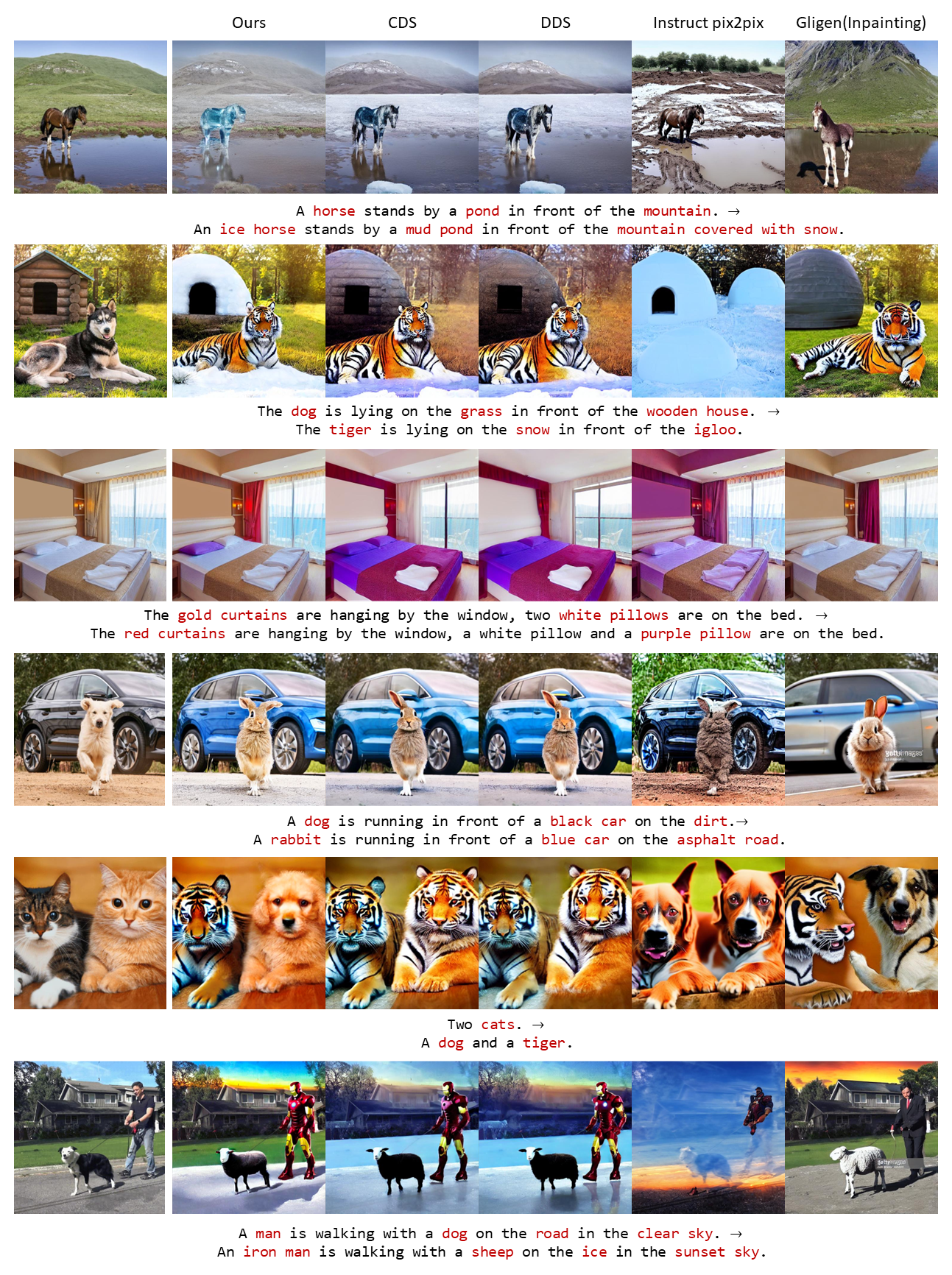}
	\caption{
The benchmark result of Ground-A-Score with other baseline models using the same editing prompts. 
}
	\label{fig_result_benchmark}
\end{figure}

\section{Experimental Results}

\subsubsection{Experimental Settings}
We used StableDiffusion 1.5~\cite{stablediffusion} as the base T2I diffusion model and GPT4-vision~\cite{GPT4} as the multimodal LLM model. For a zero-shot grounding model, we utilized groundingDINO~\cite{grounding_dino}.
Details of the Chain-of-Thought~\cite{cot} prompt used for LLM, the editing prompts, and the hyperparameters that we've used are included in the Supplementary Material. 
We will release the source code upon acceptance of the paper.

\subsubsection{Qualitative Results}
We compared the outcomes from Ground-A-Score and other conventional image editing techniques with various case studies to compare their behavior and relative performance. 
For this experiment, CDS\cite{cds} and DDS~\cite{dds} used only full prompts of source and target. In the case of InstructPix2Pix~\cite{instructpix2pix}, we followed the authors and iteratively applied the image modification with each editing target's source prompt and target prompt, with a format of \emph{``Turn A into B''}. For GLIGEN~\cite{gligen}, we entered the full target prompts used by Ground-A-Score, the mask for each object, and the target prompt assigned to each mask. 

Figure~\ref{fig_result_benchmark} shows the output image of Ground-A-Score and various suggested baselines with common editing prompts.
While our methodology enables us to effectively change various elements within the images, CDS and DDS often cannot transform several target objects or modified regions that should be unchanged. 
InstructPix2Pix gradually lost the detailed appearance of the source image and occasionally introduced unintended changes as the process was repeated. As an inpainting-based image editing method, GLIGEN turned out to neglect the source image's character in some example scenarios.

\subsubsection{Quantitative Results}
To the best of our knowledge, there is no benchmark dataset that evaluates the image editing method's ability to edit multiple attributes individually. 
Therefore, in order to quantitatively evaluate the performance, we employed the multimodal LLM again to recommend possible editing targets in the given image. The mLLM was asked to find three elements to modify and express those changes into a generated source and target prompts. Then, we put those prompts and the source image into our pipeline and the other baseline methods. The source images were prepared by randomly sampling 300 images from the Visual Genome dataset~\cite{visualgenome}. 
We compared Ground-A-Score with the other image editing methods using a CLIP~\cite{clip} cosine similarity score with CLIP(ViT-B/16) and LPIPS~\cite{lpips} perceptual loss as an image-text alignment and image quality metric, respectively.

\begin{table}[tb]  
  \caption{Quantitative comparison for multi-attribute editing with randomly sampled source images from Visual Genome dataset~\cite{visualgenome} and synthesized editing prompts. The best performance for each metric is written in bold.}
  \label{tab:headings2}
  \centering
  \resizebox{.7\textwidth}{!}{
  \begin{tabular}{@{}l>{\centering\arraybackslash}m{2.5cm}>{\centering\arraybackslash}m{2.5cm}>{\centering\arraybackslash}m{2cm}@{}} 
    \toprule
    & \textbf{LPIPS}[$\downarrow$] & \textbf{CLIP score}[$\uparrow$] & \textbf{CLIP score} \textbf{(masked)}[$\uparrow$] \\ 
    \midrule
    \textbf{GLIGEN}~\cite{gligen} & 0.5556 & 30.34 & 23.66 \\
    \textbf{InstructPix2Pix}~\cite{instructpix2pix} & 0.6948 & 27.47 & 23.04 \\
    \textbf{DDS}~\cite{dds} & 0.4022 & \textbf{32.33} & 24.47  \\
    \textbf{CDS}~\cite{cds} & 0.3839 & 31.93 & 24.48  \\
    \textbf{Ours} & \textbf{0.3668} & 30.49 & \textbf{25.07}  \\
    \bottomrule
  \end{tabular}}
\end{table}

Table~\ref{tab:headings2} contains the image-text alignment and the image quality of the edited image by Ground-A-Score and other baseline models. Ground-A-Score achieved a better image quality with small LPIPS conceptual loss compared to the other image editing methods. 
Although the CLIP score of Ground-A-Score was slightly lower than the other distillation-based methods when the cosine similarity was calculated with the entire output image and a full target prompt, this happened because DDS and CDS could trade off the stationary image features for the alignment; weakening the constraints of image editing, the alignment can be improved by optimizing the output beyond the context of the original image(\eg background). 
When the CLIP score was separately measured for each editing target with the corresponding regions in the output image, Ground-A-Score had the highest agreement between the prompt, as shown in the rightmost column in Table~\ref{tab:headings2}.

\begin{table}[tb]
  \caption{The average user study score of Ground-A-Score and other baseline models on three criteria.}
  \label{tab:userstudy}
  \centering
  \resizebox{.7\textwidth}{!}{
  \begin{tabular}{@{}l>{\centering\arraybackslash}m{2.5cm}>{\centering\arraybackslash}m{2.5cm}>{\centering\arraybackslash}m{2cm}@{}} 
    \toprule
    & \textbf{Fidelity}[$\uparrow$] & \textbf{Preservation}[$\uparrow$] & \textbf{Quality}[$\uparrow$]\\ 
    \midrule
    \textbf{GLIGEN}~\cite{gligen} & 2.38 & 2.72 & 3.06 \\
    \textbf{InstructPix2Pix}~\cite{instructpix2pix} & 2.03 & 2.37 & 3.08 \\
    \textbf{DDS}~\cite{dds} & 3.33 & 3.54 & 3.43 \\
    \textbf{CDS}~\cite{cds} & 3.38 & 3.75 & 3.65 \\
    \textbf{Ours} & \textbf{4.65} & \textbf{4.65} & \textbf{4.38} \\
    \bottomrule
  \end{tabular}}
\end{table}

Furthermore, to compare the model's performance to provide the output image more appealing to the end user, we've additionally conducted a user survey and collected responses from 25 subjects. Each subject was asked to score the output images from Ground-A-Score and the other baseline models with three criteria: fidelity for the input prompt, preservation of the unchanged features, and the overall image quality. The average score of the user study is listed in Table~\ref{tab:userstudy}, which shows that our method received the highest average score.

Through these demonstrations, we concluded that our method most appropriately modifies the object as intended in the prompt compared to the existing image editing methods.

\subsubsection{Ablation Studies} \label{result_abl}

We've conducted ablation studies for each major component of the proposed Ground-A-Score to visualize and emphasize their role.

\begin{figure}[tb]
	\centering
 \includegraphics[width=\textwidth]{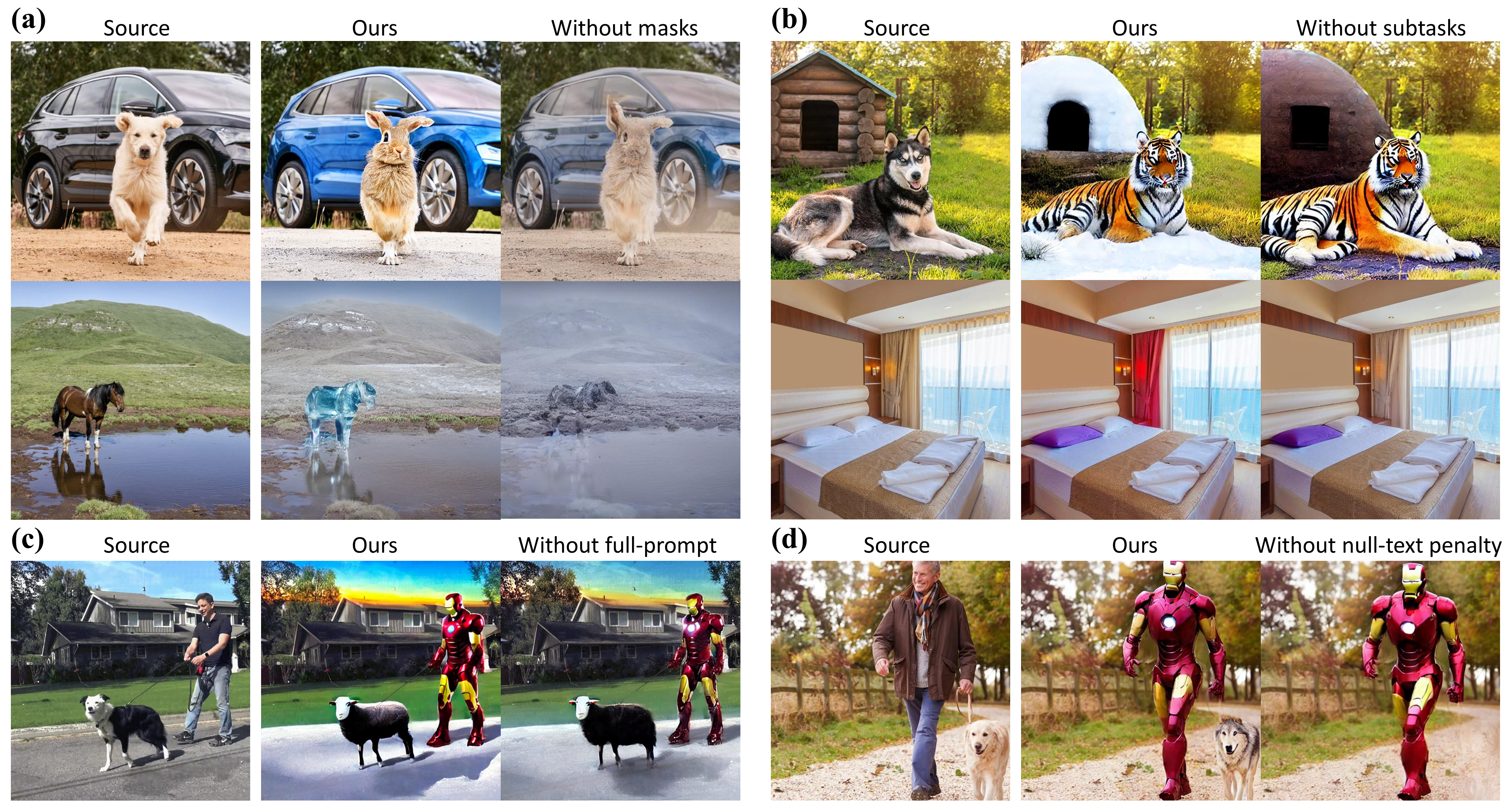}
	\caption{
Ablation study on the components of the proposed Ground-A-Score: object masking, subtask scheduling, full-prompt editing, and null-text penalty.
}
	\label{fig_result_ablation}
\end{figure}

Figure ~\ref{fig_result_ablation}-(a) showed the output image was significantly disrupted when the gradient masking for each subtask and the full-prompt guidance was removed; each subtask affected the region unrelated to its target, resulting in a blurred background and an incapability to perform the intended changes. 
Removing the subtask scheduling is equivalent to solely utilizing full-prompt guidance. This turned out to make the result fail to reflect some of the total editing prompts, even when we applied a total mask $M$ to guide the change only happens on the target objects. 
Figure~\ref{fig_result_ablation}-(b) shows that the absence of the subtasks fails to turn the wooden house into an igloo, turn the grass into snow, and turn the curtains red.

When we didn't include the full-prompt guidance term, the boundaries of the overlapping attributes were often blurred and distorted, such as the edges between Iron Man, sheep, and the roads in Figure~\ref{fig_result_ablation}-(c). This suggests that the full-prompt guidance serves to revise the oddness of the image when it is edited.

Figure~\ref{fig_result_ablation}-(d) depicts an example where the null-text penalty aids the output image to convert the editing object to the target condition.
When the image gradient occasionally drives the image to remove the object, the null-text penalty degrades this effect by multiplying a small coefficient by the gradient. Otherwise, the object couldn't receive guidance towards the target condition and eventually disappeared.

\section{Conclusion}

In this work, we focused on tackling the challenging problem of simultaneously changing multiple elements in a single image at once. To effectively resolve this problem, we applied the principle of divide-and-conquer to the conventional distillation-based image editing methods.
In the process, we suggested an effective gradient aggregation method and regularization to overcome the possible malfunctions.
We also combined this method with a zero-shot grounding model and a multi-model LLM to create a pipeline to easily edit multiple attributes. 

We believe the proposed strategy of subtask combination with suitable regularization is a general approach that is easily applicable to other diffusion model-based image editing methodologies including PNP~\cite{plug_and_play}, P2P~\cite{prompt_to_prompt}, and MasaCtrl~\cite{masactrl}, and has the potential to improve their multi-attribute capability.

\newcolumntype{L}[1]{>{\raggedright\let\newline\\\arraybackslash\hspace{0pt}}m{#1}}
\newcolumntype{C}[1]{>{\centering\let\newline\\\arraybackslash\hspace{0pt}}m{#1}}
\newcolumntype{R}[1]{>{\raggedleft\let\newline\\\arraybackslash\hspace{0pt}}m{#1}}

\title{Supplementary Materials for Ground-A-Score: Scaling Up the Score Distillation for Multi-Attribute Editing} 

\title{Ground-A-Score} 

\author{Hangeol Chang\inst{*} \and
Jinho Chang\inst{*} \and
Jong Chul Ye\inst{}}

\authorrunning{Hangeol Chang, Jinho Chang, and Jong Chul Ye.}

\institute{Kim Jaechul Graduate School of AI, KAIST  \\
\email{\{hangeol, jinhojsk515, jong.ye\}@kaist.ac.kr}\\
* Equal contribution}
\maketitle

\section{Experimental Details}  \label{section_details}
\subsection{Image Optimization Process}
For the qualitative analyses in the main manuscript, the image latent optimization with the gradient from Ground-A-Score was done with SGD optimizer until the image converges with a Classifier-Free Guidance weight $\omega$ of 7.5. We used the same number of optimization steps on each image for other distillation-based image editing methods~\cite{dds, cds} in the benchmark study. In the case of quantitative experiments with hundreds of images in the Visual Genome dataset~\cite{visualgenome}, we fixed the number of optimization steps to 500. 

Zero-shot grounding for mask preparation was done by groundingDINO~\cite{grounding_dino}, where we used the mask with the highest probability for each given phrase. If the same phrase appears more than once for a single image, the n most probable masks were used.

When the mask for a subtask is overlapped by another subtask's mask, we multiply a factor of 0.3 to the image gradient from the larger mask's overlapping region.

\subsection{Full-Prompt Guidance}

For the full-prompt guidance hyperparameter $\alpha$, we used a different value for each optimization step depending on the sampled t, the T2I diffusion model's noise scheduling timestep. We observed that full-prompt guidance with the same $\alpha$ changes the output image more drastically when the noised latent's Signal-to-Noise Ratio (SNR) is small. Therefore, we divided the overall noise scheduling timestep of StableDiffusion into five equal ranges and assigned the $\alpha$ value from 0.5 to 0.1 as the SNR decreases.

\subsection{Null-Text Penalty}
\begin{figure}[ht!]
	\centering
 \includegraphics[width=\textwidth]{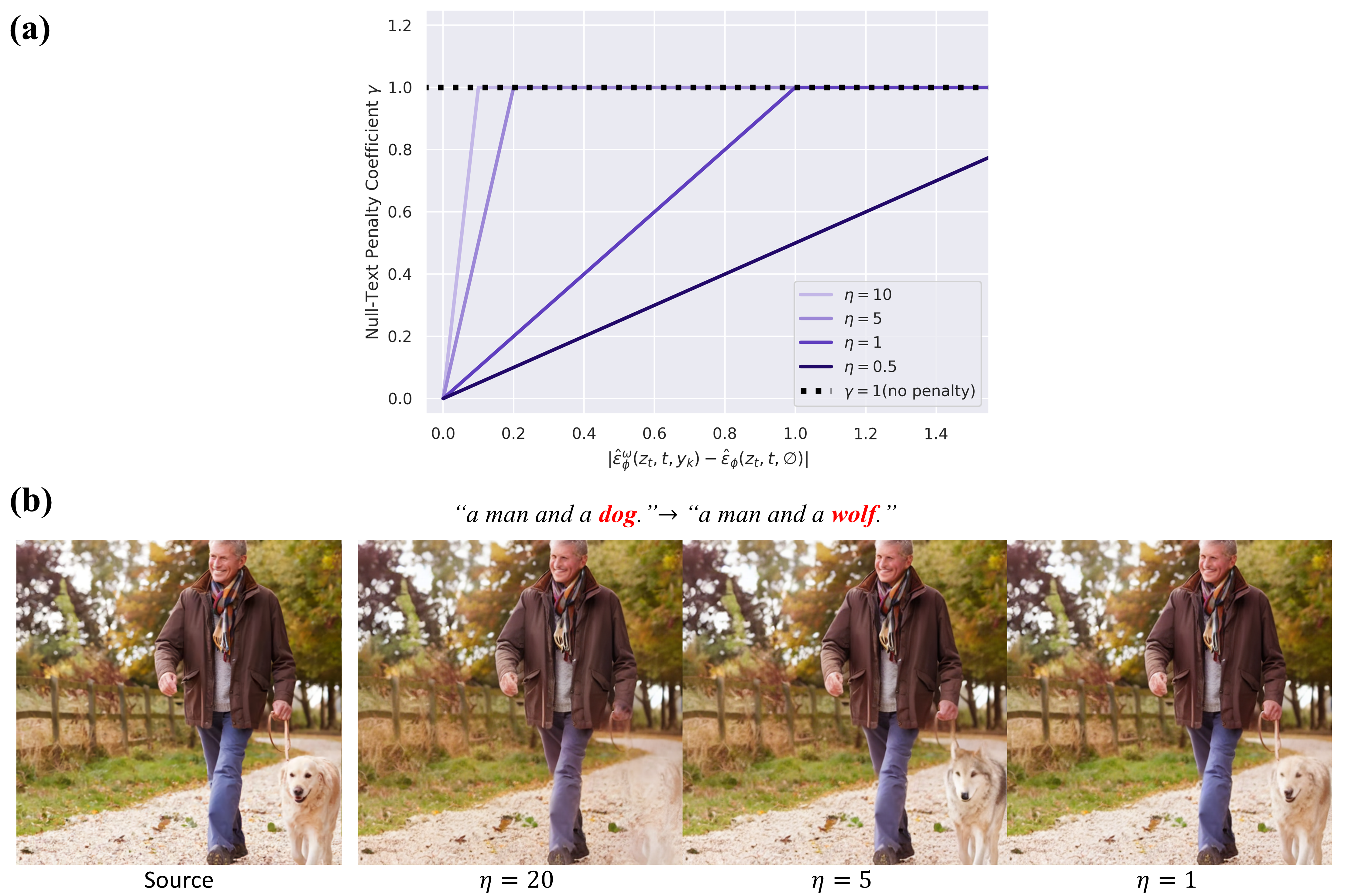}
	\caption{
The effect of changing hyperparameter $\eta$ for the null-text penalty. (a) The plot of the null-text penalty coefficient $\gamma$ against the difference between $\hat{\epsilon}^{\omega}_{\phi} (z_t,t,y_k)$ and $\hat{\epsilon}_{\phi} (z_t,t,\varnothing)$ with different $\eta$. The dotted line of $\gamma=1$ represents the scenario without a null-text penalty. (b) Ground-A-Score image editing outputs using three different $\eta$ with a shared source image and editing prompt. The images were optimized with the same number of optimization steps and the optimization step size.
}
	\label{fig_ntp}
\end{figure}

The $\eta$ in the null-text penalty term determines the range of the computed gradient being penalized by changing the penalty coefficient $\gamma$. As shown in Figure~\ref{fig_ntp}-(a), when $\eta$ becomes infinitely large, most of the calculated gradients are unchanged by getting multiplied by 1. This would defeat the purpose of introducing the null-text penalty and fail to prevent the disappearance of certain objects in the image as in the example of $\eta=20$ in Figure~\ref{fig_ntp}-(b).
On the other hand, as $\eta$ goes to 0, more portion of the calculated gradients are penalized with a small $\gamma$. Although this wouldn't introduce a wrong optimization direction and corrupt the image, it significantly slows the overall image transformation, as in the example of $\eta=1$ in Figure~\ref{fig_ntp}-(b). 

We empirically found that $\eta=5$ works well for the overall images we've tested, which is the value we used for the experiments.
Also, assuming that the null-text penalty is not needed for large and prominent objects, we only applied the null-text penalty to subtasks whose corresponding mask's size is less than 15\% of the whole image size to speed up the image optimization.

\section{Additional Results}
\begin{figure}[ht!]
	\centering
 \includegraphics[width=\textwidth]{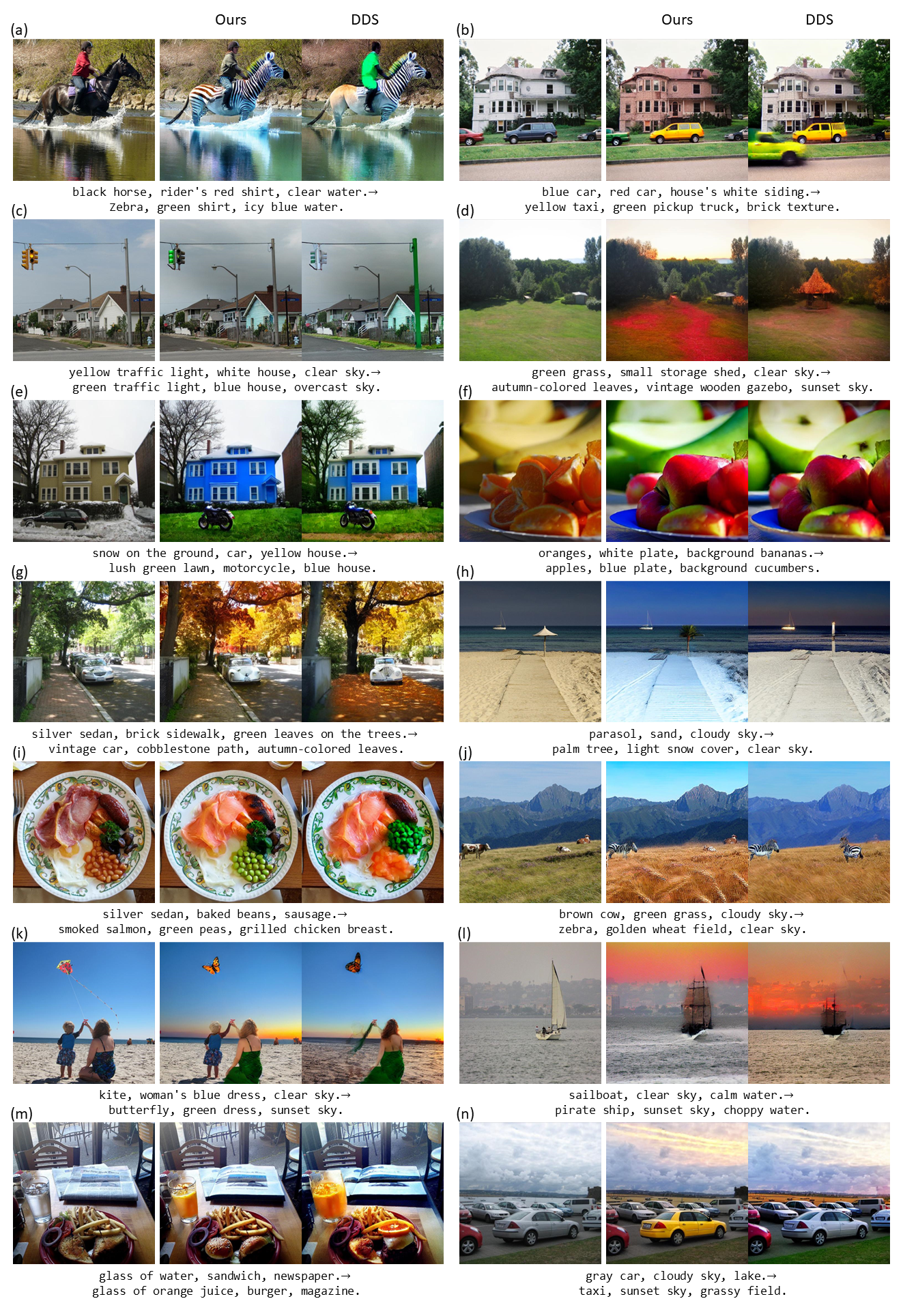}
	\caption{
The image editing results from Ground-A-Score and DDS, with synthetic image editing scenarios and editing prompts.
}
	\label{fig_examples}
\end{figure}

Among the Ground-A-Score output images of automatically generated editing scenarios by prompts in Table~\ref{tab:llm_prompt_recommend} and Table~\ref{tab:llm_prompt_editing_new}, we included 14 images with high Masked CLIP cosine similarity scores in Figure~\ref{fig_examples}. The phrases in Figure~\ref{fig_examples} under each output image are the prompts used for editing without articles. The optimization process followed Section~\ref{section_details}, the same as the quantitative experiments conducted in the main manuscript.
  
In all cases, Ground-A-Score transformed the target objects perceptually more accurately compared to DDS~\cite{dds}. However, when we calculated those images' CLIP score with the full-sentence prompt and the entire image, we found that DDS still had a higher CLIP score as shown in Table~\ref{tab:numbers}, which also happened in the main results. 
As we discussed in the main manuscript, this is likely due to the effect of DDS trading off the preservation of the source image to increase the CLIP score, and the inability of the CLIP score to accurately interpret long sentences.
For example, in DDS's output, the green pickup truck in Figure~\ref{fig_examples}-(b), the vintage wooden gazebo in Figure~\ref{fig_examples}-(d), and the orange juice in Figure~\ref{fig_examples}-(m) became bigger, more emphasized, and even appeared in a different location compared to the corresponding object in the source image. 
On the contrary, in the case of the masked CLIP score, which is the mean CLIP score between each subtask and the image's region in the corresponding mask, the outputs from Ground-A-Score achieved a higher similarity. With this result, we insist that the masked CLIP score is a more appropriate metric for evaluating multi-attribute editing performance than a vanilla CLIP score.

\begin{table}[tb]  
  \caption{Quantitative results from the image editing results in Figure~\ref{fig_examples}.}
  \label{tab:numbers}
  \centering
  \resizebox{.7\textwidth}{!}{
  \begin{tabular}{@{}l>{\centering\arraybackslash}m{2.5cm}>{\centering\arraybackslash}m{2.5cm}>{\centering\arraybackslash}m{2cm}@{}} 
    \toprule
    & \textbf{LPIPS}[$\downarrow$] & \textbf{CLIP score}[$\uparrow$] & \textbf{CLIP score} \textbf{(masked)}[$\uparrow$] \\ 
    \midrule
    \textbf{DDS}~\cite{dds} &39.44 & \textbf{35.56} & 25.90  \\
    \textbf{Ours} & \textbf{39.33} & 35.51 & \textbf{27.61}  \\
    \bottomrule
  \end{tabular}}
\end{table}

\section{Detailed Editing Prompts.}

Table~\ref{tab:prompt_fig1_1} contains the full-sentence editing prompts $\{X, Y\}$ and the subtasks $\{x_k, y_k\}$ for each image editing scenario illustrated in the main manuscript. These prompts were obtained with GPT4-vision~\cite{GPT4}, using a LLM prompt described in Section~\ref{section_4_2}.

\begin{table}[!ht]
    \centering
    \caption{Full-sentence editing prompts and the corresponding subtasks for each image editing scenario in the main manuscript figures.}
    {\ssmall
    \begin{tabular}{C{0.08\textwidth}|L{0.92\textwidth}} \toprule
    \multicolumn{2}{c}{\textbf{Figure 1-1}} \\ \midrule
    $X$ & A teddy bear is sitting in front of a stone structure with a clear sky above and green grass below. \\ 
    $Y$ & A classic Greek vase is sitting in front of a stone structure with a sunset sky above and autumn-colored leaves below. \\ 
    $\{x_1,y_1\}$ & \{a teddy bear, a classic greek vase\} \\
    $\{x_2,y_2\}$ & \{clear sky, sunset sky\} \\ 
    $\{x_3,y_3\}$ & \{green grass, autumn-colored leaves\} \\ \midrule 
    
    \multicolumn{2}{c}{\textbf{Figure 1-2, Figure 4-2, Figure 5-(b)-1}} \\ \midrule
    $X$ & the dog is lying on the grass in front of the wooden house. \\ 
    $Y$ & the tiger is lying on the snow in front of the igloo. \\ 
    $\{x_1,y_1\}$ & \{the dog, the tiger on the snow\} \\ 
    $\{x_2,y_2\}$ & \{the grass, the snow\} \\ 
    $\{x_3,y_3\}$ & \{the wooden house, the igloo\} 
    \\ \midrule 

    \multicolumn{2}{c}{\textbf{Figure 1-3, Figure 4-5}} \\ \midrule
    $X$ & two cats. \\ 
    $Y$ & a dog and a tiger. \\ 
    $\{x_1,y_1\}$ & \{a cat, a dog\} \\ 
    $\{x_2,y_2\}$ & \{a cat, a tiger\} \\ \midrule 

    \multicolumn{2}{c}{\textbf{Figure 1-4, Figure 4-4, Figure 5-(a)-1}} \\ \midrule
    $X$ & A dog is running in front of a black car on the dirt. \\ 
    $Y$ & A rabbit is running in front of a blue car on the asphalt road. \\ 
    $\{x_1,y_1\}$ & \{a dog, a cat\} \\ 
    $\{x_2,y_2\}$ & \{a black car, a blue car\} \\ 
    $\{x_3,y_3\}$ & \{the dirt, the asphalt road\} \\ \midrule 

    \multicolumn{2}{c}{\textbf{Figure 1-5}} \\ \midrule
    $X$ & Mashed potatoes are on a plate next to a steak and broccoli. \\ 
    $Y$ & A green salad is on a plate next to grilled salmon and parsley. \\ 
    $\{x_1,y_1\}$ & \{mashed potatoes, a green salad\} \\ 
    $\{x_2,y_2\}$ & \{a steak, grilled salmon\} \\ 
    $\{x_3,y_3\}$ & \{broccoli, parsley\} \\ \midrule 
    
    \multicolumn{2}{c}{\textbf{Figure 1-6, Figure 4-6, Figure 5-(c)}} \\ \midrule
    $X$ & a man is walking with a dog on the road in the clear sky. \\ 
    $Y$ & an iron man is walking with a sheep on the snow road in the sunset sky. \\ 
    $\{x_1,y_1\}$ & \{a man, an iron man\} \\ 
    $\{x_2,y_2\}$ & \{a dog, a sheep\} \\ 
    $\{x_3,y_3\}$ & \{the road, the snow road\} \\ 
    $\{x_4,y_4\}$ & \{the clear sky, the sunset sky\} \\ \midrule 

    \multicolumn{2}{c}{\textbf{Figure 4-1, Figure 5-(a)-2}} \\ \midrule
    $X$ & a horse stands by a pond in front of the mountain. \\ 
    $Y$ & an ice horse stands by a mud pond in front of the mountain covered with snow. \\ 
    $\{x_1,y_1\}$ & \{a horse and reflection, an ice and reflection\} \\ 
    $\{x_2,y_2\}$ & \{a pond, a mud\} \\ 
    $\{x_3,y_3\}$ & \{the mountain, the mountain slightly covered with snow\} \\ \midrule 

    \multicolumn{2}{c}{\textbf{Figure 4-3, Figure 5-(b)-2}} \\ \midrule
    $X$ & The gold curtains are hanging by the window, two white pillows are on the bed. \\ 
    $Y$ & The red curtains are hanging by the window, a white pillow and a purple pillow are on the bed. \\ 
    $\{x_1,y_1\}$ & \{The gold curtains, The red curtains\} \\ 
    $\{x_2,y_2\}$ & \{a white pillow, a purple pillow\} \\ \midrule 

    \multicolumn{2}{c}{\textbf{Figure 5-(d)}} \\ \midrule
    $X$ & A middle aged man walks with the dog. \\ 
    $Y$ & The iron man walks with the wolf. \\ 
    $\{x_1,y_1\}$ & \{a middle aged man, the iron man\} \\ 
    $\{x_2,y_2\}$ & \{the dog, the wolf\} \\ 
    \bottomrule
    \end{tabular}
    }
    \label{tab:prompt_fig1_1}
\end{table}

\section{Multimodal LLM prompts}
\subsection{Prompt for Synthetic Editing Task Generation.}
Table~\ref{tab:llm_prompt_recommend} shows the prompt entered in GPT4-vision that we've utilized to generate synthetic image editing scenarios for a large number of images in the Visual Genome~\cite{visualgenome} dataset.

\begin{table}[!ht]
    \centering
    \caption{The multimodal LLM prompt to generate a synthesized image editing task for a given image, which we used for a benchmark study on the images from Visual Genome dataset.}
    {\scriptsize
    \begin{tabular}{L{\textwidth}} \toprule 
    As a researcher in image editing, your task is to review a provided photograph, identify three distinct elements within the image that could be modified using image editing techniques, and suggest three modifications. \\
    Your suggestions should focus on altering tangible elements to transform the visual experience. \\ 
    \\
    For example:\\ 
    -If there is an animal present, suggest replacing it with a different animal (e.g., change a cat into a dog).\\
    -If a particular object's color stands out, propose altering its color (e.g., change a black car into a red car).\\
    -Consider transforming human figures into recognizable fictional characters (e.g., change a man into Iron Man).\\
    -Additionally, you may suggest changes to the environment (e.g., change grass into snow) or the atmosphere (e.g., change a clear sky into a cloudy sky).\\ 
    \\
    However, ensure that each modification:\\
    -Replace one specific, tangible element with another. For example, instead of merely saying 'make the sky cloudy,' specify 'change the clear sky into the cloudy sky.'\\
    -Avoid overlapping with other elements.\\
    -Avoid altering elements that dominate the entire image.\\
    -Select objects of medium size for modification. Editing objects that are either too small or too large is not recommended.\\
    -Ensures the transformation maintains a similar form for simplicity.\\
    -Don't use the format `change X, A into B', use `change A into B'.\\
    -Don't ask to change letters.\\ 
    \\
    Format your suggestions as follows: change = [change A into B. change C into D. change E into F.]\\
    \{\texttt{input source image}\} \\
        \bottomrule
    \end{tabular}
    }
    \label{tab:llm_prompt_recommend}
\end{table}

\subsection{Prompt for Full-Sentence Prompt and Subtasks}   \label{section_4_2}
Table~\ref{tab:llm_prompt_editing_new} includes the GPT4-vision input prompt to generate full-text prompts and the corresponding subtasks for a given image and the editing request from the user.

\begin{table}[!ht]
    \centering
    \caption{The Chain-of-Thought multimodal LLM prompt to schedule subtasks for a given full-text image editing query.}
    {\tiny
    \begin{tabular}{L{\textwidth}} \toprule 
    When you receive a description of how to change a given image and image, create sentences that represent the current and desired states of the picture. To do this, you use two arrays: ‘source\_list’ and ‘target\_list’. Each array contains elements representing the objects or features to be changed, as many as requested, followed by a summarizing sentence of the scene. The list order should match the sequence of transformation requests. If objects are closely related (e.g., an animal lying on furniture), their relationship should be explicitly mentioned, even if it is not the last summarized sentence, ensuring the context is maintained throughout the transformation. When occurring the unnaturally severe change of an object’s material, only the material of the object is entered in ‘target\_list’. full form has to be expressed only in the last sentence. 
    Additionally, you must decide whether each element’s original shape is preserved during the transformation. This
    decision is captured in the ‘preserve\_form’ array, where ’1’ indicates shape preservation, and ’0’
    indicates that the shape can change. The last number in this array reflects the overall preservation
    of form in the scene. The Preserve form uses a lot of resources, so if you’re not sure, leave it at 0.
    \\\\
    \textbf{Example 1}:\\
    \textbf{Question}:
    Apply the rules explained above. an image of a dog running in front of a car on the dirt is given.
    How should the ‘source\_list’, ‘target\_list’, and ‘preserve\_form’ arrays be structured?\\
    Requests: Change a dog into a cat. Change a car into a Lego car. Change the dirt into the asphalt road.\\
    \textbf{Explanation}: 
    The things requested to be changed in the image are a dog, a car and the dirt in the order requested. Image can be described with the things as "a dog running in front of a car on the dirt.". So, the 'source\_list' is ["a dog," "a car," "the dirt," "a dog running in front of a car on the dirt"]. The requested changes are to change a dog into a cat, a car into a Lego car, and the dirt into the asphalt road. Therefore, the 'target\_list' is ["a cat," "a Lego car," "the asphalt road," "A cat is running in front of a Lego car on the asphalt road."].
    The dog and the cat can take similar overall poses, so it is more natural to maintain the shape when converting a dog into a cat. The car’s form changes significantly as it becomes a Lego car. Asphalt and dirt have similar shapes, but since their shapes are simple, there is no need to force them to maintain their shape. Lastly, since the appearance of the car will change significantly, it is better not to force it to maintain the entire scene. Therefore, the 'preserve\_form' is [1,0,0,0].\\
    \textbf{Final answer}:\\
    'source\_list': ["a dog", "a car", "the dirt", "A dog is running in front of a car on the dirt."]\\
    'target\_list': ["a cat", "a Lego car", "the asphalt road", "A cat is running in front of a Lego car on the asphalt road."]\\
    'preserve\_form': [1, 0, 0, 0]\\
    \\
    \textbf{Example 2}: \\
    \textbf{Question}:
    Apply the rules explained above. an image of a dog lying on the grass is given.
    How should the 'source\_list', 'target\_list', and 'preserve\_form' arrays be structured?\\
    Requests: change a dog into a tiger. Change the grass into the snow.\\
    \textbf{Explanation}: 
    The things requested to be changed in the image are a dog and the grass. The requested changes are to change a dog into a tiger and the grass into the snow. The transformation involves changing both the animal and the surface it lies on, maintaining a close relationship between the two. Since the dog and the grass are in close contact over a long area, the dog and grass, tiger and snow should express together. So, the 'source\_list' is ["the dog on the grass", "the grass", "The dog is lying on the grass."] and 'target\_list' is ["the tiger on the snow", "the snow", "The tiger is lying on the snow."].
    When changing a dog into a tiger, it is natural to maintain a lying position. When changing grass into
    snow, shape may change because It can change from a sharp grass shape to a smooth snowy shape.
    Cases where the shape can change are included, so the entire scene can also change shape. Therefore, the 'preserve\_form' is [1,0,0].\\
    \textbf{Final answer}:\\
    'source\_list': ["the dog on the grass", "the grass", "The dog is lying on the grass."]\\
    'target\_list': ["the tiger on the snow", "the snow", "The tiger is lying on the snow."]\\
    'preserve\_form': [1, 0, 0]\\
    \\
    \textbf{Example 3}: \\
    \textbf{Question}:
    Apply the rules explained above. an image of a horse is standing near pond with its reflection visible, with the mountain
    in the background is given.\\
    Requests: Change horse and its reflection into ice horse and its reflection. Change the mountain into snowy mountain.\\
    \textbf{Explanation}: 
    The things requested to be changed in the image are horse and its reflection, and the mountain. Image can be described with the things as "A horse is standing near pond with its
    reflection visible, with a mountain in the background." . So, the 'source\_list' is ["a horse and its reflection", "a mountain", "A horse is standing near pond with its reflection visible, with a mountain in the background."]
    The requested changes are to change horse and its reflection into ice horse and its reflection and the mountain into snowy mountain. Because a severe change in material occurred from horse to ice horse, only material should be written in the 'target\_list'. So 'target\_list' is ["an ice and its reflection", "a snowy mountain", "Ice horse is standing near pond with its reflection visible, with a snowy mountain in the background."]
    Since the shape of the horse must be maintained, the first term of 'preserve\_form' is 1. 
    As snow accumulates, changes in shape may occur, so the second term is 0. Because the case
    where the shape can change is included, the third term is 0. Therefore, the 'preserve\_form' is [1,0,0].\\
    
    \textbf{Final answer}:\\
    'source\_list': ["a horse and its reflection", "a mountain", "A horse is standing near pond with its\\
    reflection visible, with a mountain in the background."]
    'target\_list': ["an ice and its reflection", "a snowy mountain", "Ice horse is standing near pond with its\\
    reflection visible, with a snowy mountain in the background."].
    'preserve\_form': [1, 0, 0]\\
    \\
    \textbf{Question}: Apply the rules explained above. Please refer to the picture and request below.
    How should the 'source\_list', 'target\_list', and 'preserve\_form' arrays be structured?\\
    Requests:\{responses\}\\
    Answer in the form of\\
    Explanation: ...\\
    Final answer: ...\\
    as in the examples above.\\
    \{\texttt{input source image}\} \\
    \bottomrule
    \end{tabular}
    }
    \label{tab:llm_prompt_editing_new}
\end{table}

\section{Code Availability}
The official implementation of Ground-A-Score is available at the anonymous GitHub repository, \href{https://github.com/Ground-A-Score/Ground-A-Score/}{github.com/Ground-A-Score/Ground-A-Score}, which includes the script for the input preparation pipeline and image optimization.

\clearpage  

%
%

\clearpage  
\bibliographystyle{splncs04}
\bibliography{main}
\end{document}